\documentclass[letterpaper, 10 pt, conference]{ieeeconf}  

\IEEEoverridecommandlockouts                              

\usepackage[table,xcdraw]{xcolor}
\usepackage{multirow}
\usepackage{booktabs}
\usepackage{bm}
\usepackage{amsmath}
\usepackage{amssymb}
\usepackage{graphicx}
\usepackage{array}
\usepackage{mathtools}
\usepackage{hyperref}
\hypersetup{
    colorlinks=true,  
    linkcolor=black,  
    citecolor=black,  
    filecolor=black,  
    urlcolor=blue     
}
\usepackage{float}
\usepackage[final]{changes}
\definechangesauthor[name={deb}, color=red]{d}

\makeatletter
\def\endtable{\end@float}
\def\endfigure{\end@float}
\makeatother
\usepackage[caption=false, font=footnotesize]{subfig}
\usepackage{arydshln}
\usepackage{booktabs}

\usepackage[nolist,nohyperlinks]{acronym}
\newacro{DDPM}[DDPM]{Denoising Diffusion Probabilistic Models}
\newacro{DDBM}[DDBM]{Denoising Diffusion Bridge Models}
\newacro{CFM}[CFM]{Conditional Flow Matching}
\newacro{CFG}[CFG]{Classifier Free Guidance}
\newacro{NOMAD}[NoMaD]{Navigation with Goal Masked Diffusion}
\newacro{FLOWNAV}[FlowNav]{FlowNav}
\newacro{SCALENET}[MetricNet]{MetricNet}
\newacro{ours}[MetricNav]{MetricNav}
\newacro{CLS}[CLS]{Classifier Token}
\newacro{MSE}[MSE]{Mean Squared Error}
\newacro{nomad}[NoMaD]{NoMaD}
\newacro{SLAM}[SLAM]{Simultenous Localization And Mapping}
\newacro{TSDF}[TSDF]{Truncated Signed Distance Function}

\iffalse 
  \newcommand{\gode}[1]{\noindent}
  \newcommand{\nayak}[1]{\noindent}
  \newcommand{\oliveira}[1]{\noindent}
\else
  \newcommand{\gode}[1]{\textcolor{red}{\bf [SG: #1]}}
  \newcommand{\nayak}[1]{\textcolor{green}{\bf [AN: #1]}}
  \newcommand{\oliveira}[1]{\textcolor{magenta}{\bf [DO: #1]}}

\fi

\title{\LARGE \bf
MetricNet: Recovering Metric Scale in Generative Navigation Policies
}

\author{Abhijeet Nayak*$^{1}$, Débora Oliveira Makowski*$^{1}$, Samiran Gode*$^{1}$, Cordelia Schmid$^{2}$ and Wolfram Burgard$^{1}$
\thanks{${1}$ Artificial Intelligence and Robotics Lab, Department of Computer Science and Artificial Intelligence, University of Technology Nuremberg, Germany. \{\tt\small\textit{firstname.lastname}\}\tt\small{@utn.de}}%
\thanks{${2}$ Inria, Ecole Normale Supérieure, CNRS, PSL Research University, France. \{\tt\small\textit{firstname.lastname}\}\tt\small{@inria.fr}}%
\thanks{* These authors contributed equally to the work. Listing order is random.}%
\thanks{Available at \url{https://utn-air.github.io/metricnet}}%
}

\begin{document}

\maketitle
\thispagestyle{empty}
\pagestyle{empty}

\begin{abstract}
Generative navigation policies have made rapid progress in improving end-to-end learned navigation.
Despite their promising results, this paradigm has two structural problems.
First, the sampled trajectories exist in an abstract, unscaled space without metric grounding. 
Second, the control strategy discards the full path, instead moving directly towards a single waypoint. 
This leads to short-sighted and unsafe actions, moving the robot towards obstacles that a complete and correctly scaled path would circumvent.
To address these issues, we propose \acs{SCALENET}, an effective add-on for generative navigation that predicts the metric distance between waypoints, grounding policy outputs in \replaced[id=d]{metric}{real-world} coordinates.
We evaluate our method in simulation with a new benchmarking framework and show that executing \acs{SCALENET}-scaled waypoints significantly improves both navigation and exploration performance.
Beyond simulation, we further validate our approach in real-world experiments.
Finally, we propose \acs{ours}, which integrates \acs{SCALENET} into a navigation policy to guide the robot away from obstacles while still moving towards the goal.
\end{abstract}
\section{Introduction}
Classical navigation methods~\cite{Lat91Rob,choset05,lavalle06} have a long history of success in complex scenarios~\cite{Bur99Exp,thrun06stanley,tranzatto2022cerberus}.
These methods reliably find paths to a goal by either generating or using pre-built maps of the environment~\cite{thrun2005probabilistic,siegwart2011introduction}. 
However, their primary drawback is the extensive calibration required to generalize to new environments or robot embodiments~\cite{cadena2016past}.
To bridge the domain gap, generative methods train visual-navigation foundation models using large-scale datasets~\cite{shah2023vint,2024flownav,cai2025navdp,sridhar2024nomad}. These methods often frame navigation as an end-to-end problem, where a diffusion-based policy~\cite{chi2023diffusion} directly predicts a sequence of 2D waypoints using the current visual observation.

Such navigation policies operate on abstract 2D waypoints.
This abstraction enables generality, allowing for cross-embodiment and scene invariance as long as the predicted direction is correct. 
The only tuning required to deploy this policy on a new robot is
the distance the agent moves towards the chosen waypoint per time step, which is generally set as $v_{\textit{max}}/f$, where $v_{\textit{max}}$ is the maximum linear velocity and $f$ is the control frequency. 
For successful deployment, the policy must generate directionally accurate waypoints at a high frequency to enable reactive control.

\begin{figure}[!t]
    \centering
    \includegraphics[width=\linewidth]{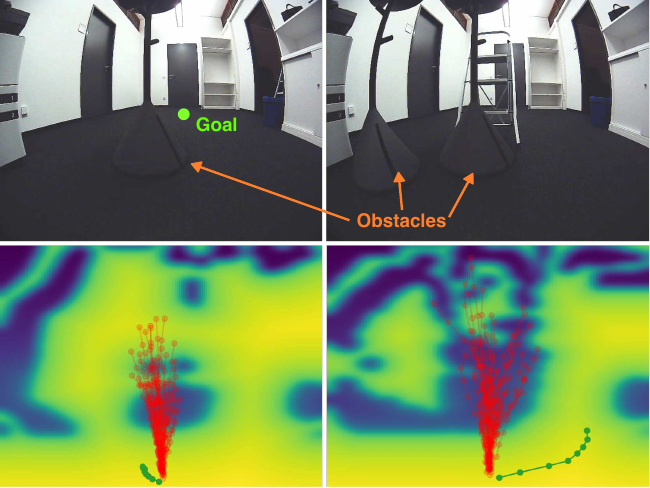}
    \caption{
    \added[id=d]{Two examples of MetricNet} \replaced[id=d]{t}{T}ransforming waypoints \deleted[id=d]{(red)} sampled from \acs{NOMAD}~\cite{sridhar2024nomad} \added[id=d]{(red)} into a collision-free plan (green) around an obstacle to reach the goal.
    First, our novel \acs{SCALENET} module grounds the waypoints in the \replaced[id=d]{3D}{real} world by predicting a metric scale. 
    Second, \acs{ours} leverages these scaled points for goal following and collision avoidance around objects.
    The trajectories are plotted on a Truncated Signed Distance Function (TSDF).
    }
    \label{fig:coverboy} 
\end{figure}

\begin{figure} 
\vspace{5.1pt}
    \centering
  \subfloat[
    Using different waypoints as goal for the velocity controller.
    \label{fig:velocity-collision}
  ]{%
    \begin{minipage}[b]{.45\linewidth}
      \centering
      \includegraphics[width=.80\linewidth]{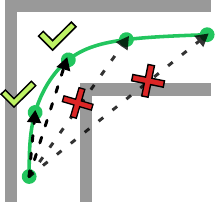}
    \end{minipage}%
  }\hfill
  \subfloat[
    Scaling the same trajectory with different scales.
    \label{fig:scale-collision}
  ]{%
    \begin{minipage}[b]{.45\linewidth}
      \centering
      \includegraphics[width=\linewidth]{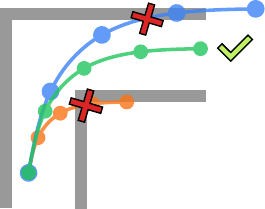}
    \end{minipage}%
  }
  \caption{
    In (a), we illustrate that when using a velocity controller, selecting only the goal waypoint can still lead to collisions despite the predicted trajectory in \replaced[id=d]{metric}{real-world} scale being free of obstacles.
    \added{Note this waypoint only points towards the direction.}
    In (b), we show that depending on the scale factor $\phi$, the resulting path may collide with walls if the scale is too small or too large for the environment. 
    This highlights the need not only for the correct \deleted[id=d]{real-world} scale but also for a controller that executes the whole trajectory in \replaced[id=d]{metric}{real-world} space.
  }
  \label{fig:colision}
\end{figure}

Despite their significant progress, the applied $v_{\textit{max}}/f$ scale does not ground the trajectory in \replaced[id=d]{metric}{real-world} coordinates.
Additionally, even though navigation policies predict a full trajectory, the agent ultimately executes only one short-term action based on a single waypoint.
This approach contrasts with the original use of diffusion policies in manipulation~\cite{chi2023diffusion}, where the full metric trajectory is executed sequentially and policies are often scene- or embodiment-specific. 
This leads to unsafe behaviors, such as moving directly towards an obstacle that the full predicted path would have navigated around, as shown in Fig.~\ref{fig:colision}. 
By grounding the trajectory in real-world coordinates, the metric representation enables explicit safety verification and goal-oriented guidance for downstream motion planners.

Our method \acs{SCALENET} addresses these problems by predicting the real distance between waypoints, enabling the robot to act directly on metric-grounded trajectories.
\acs{SCALENET} encodes the waypoints predicted from a diffusion policy and the current observation into a predicted real-world scale.
We show that acting directly on the metric waypoints outperforms the state-of-the-art control approach in both simulation and real-world deployment. 
We also show that using the grounded waypoints helps guide the trajectories away from obstacles and towards the goal, as in Fig.~\ref{fig:coverboy}.

The main contributions of our work are:
\begin{itemize}
    \item \acs{SCALENET}, a novel network that learns to predict a metric scale factor, grounding the outputs of generative policies into executable, \replaced[id=d]{metric grounded}{real-world} trajectories.
    \item A new benchmarking framework to evaluate generative visual-goal navigation and exploration policies, featuring challenging scenarios.
    \item \acs{ours}, a guidance framework that uses the metric trajectory to enforce collision avoidance and goal following simultaneously. 
    \item  An open-source release of our codebase, trained models, and real-world deployment videos. 
\end{itemize}

\section{Related Work}

\subsection{Learning-based Navigation}
Recent machine learning methods have tackled the domain generalization problem in navigation using large-scale datasets.
Approaches like RECON~\cite{shah2021rapid} and GNM~\cite{shah2023gnm} showed success with only RGB input, with the latter achieving zero-shot generalization by normalizing the action space. 
Other approaches have improved accuracy by incorporating additional sensing beyond RGB input, such as LiDAR~\cite{liang2024dtg,liang2024mtg} and monocular depth priors~\cite{2024flownav,kim2025care,zeng2025navidiffusor}.

A concurrent line of research explores language-conditioned navigation, which uses natural language to specify goals. 
For example, Poliformer~\cite{poliformer-zeng25a,hu2024flare} proposes a zero-shot discrete navigation policy through reinforcement learning. 
Parallel work StreamVLN~\cite{wei2025streamvln} introduced a dual system in which the navigation model runs faster than the language interpreter, enabling rapid reactions. 
While these advances in language instruction are promising, our primary focus is on visual navigation, where we predict actions directly from images captured by a front-facing camera.

\subsection{Generative Navigation Models}

Generative models have shown remarkable generalization capabilities in multimodal tasks such as navigation.
ViNT~\cite{shah2023vint} pioneered the use of diffusion~\cite{ho2020ddpm} to generate intermediate goal images, while NoMaD~\cite{sridhar2024nomad} applied diffusion policy~\cite{chi2023diffusion} to directly infer actions from visual observations. 
Subsequent work extended NoMaD’s diffusion framework by sampling initial noise from a learned distribution~\cite{ren2025prior,zhou2024denoising} or adopting conditional flow matching~\cite{2024flownav} for greater efficiency.
More recently, diffusion has also been used in world models for navigation~\cite{bar2025navigation}, where future visual observations are predicted from past inputs to evaluate whether a planned trajectory will successfully achieve the goal.

Although these methods demonstrate effective trajectory generation through imitation learning, they neglect the fact that training datasets typically lack collision examples. 
To address this limitation, NaviDiffusor~\cite{zeng2025navidiffusor} incorporates 3D-based obstacle avoidance into diffusion policies via gradient guidance, while NavDP~\cite{cai2025navdp} employs a critic to estimate the best collision-free path. 
Similarly, GND~\cite{liang2025gnd} encodes LiDAR scans into a traversability map to improve safety.

However, one must note that all approaches cited above rely on a large-scale dataset and therefore require input normalization for training.
At deployment, unscaling this normalization needs environment- and robot-specific fine-tuning, which may lead to collisions. 
In this work, we propose directly learning a network to project the trajectory back into \replaced[id=d]{metric}{real-world} units, thereby eliminating the need for handcrafted adjustments and enabling robust navigation.
\section{Overview}
\subsection{Problem Definition}

Generative visual-navigation models address goal-conditioned navigation by learning a policy $\pi$ which, given an observation horizon $\mathcal{O}$ and a goal image $\mathbf{g}_i$, predicts an action ${A_i = \pi(\mathbf{o}_i, \mathbf{g}_i)}$ that steers the robot towards the goal $\mathbf{g}_i$.
In the absence of a goal, $\pi$ should explore the environment without collisions.
The observation horizon $\mathcal{O}_i=[\mathbf{o}_{i}:\mathbf{o}_{i-T_O}]$ contains the images from the camera $\mathbf{o}_i$ at the timestamp $i$ and in the past horizon of length $T_O$. 
The policy $\pi$ samples from the distribution $P(A_i\mid \mathcal{O}_i)$ where $A_i$ is the set of predicted future actions in ego frame with a future horizon length $T_A$, such that $A_i=[\mathbf{a}_i:\mathbf{a}_{i+T_A}]$, where $\mathbf{a}_i$ is a single 2D waypoint.
The trajectory is only executed until a chosen waypoint $T_W \leq T_A$.
The predicted action $A_i$ is normalized by the average waypoint distance of the segment.
\replaced[id=d]{To scale it back to the metric 3D world}{To scale it back to the real world}, an unknown scale $\phi$ is needed.
In this work, we predict this scale using \acs{SCALENET}, enabling easier deployment without the need for manual tuning.

\subsection{Action Generation for Navigation Policies}

A generic navigation policy 
takes as input the observation horizon $\mathcal{O}_i$, along with an optional goal image $\mathbf{g}_i$. 
These images are first encoded into tokens, which are then processed by a transformer to create a context vector $\mathbf{c}_i$. 
This context is then passed as a condition to the generative policy to sample waypoints $A_i$ in a normalized space.. 

Previous works used different generative models to sample trajectories. 
NoMaD~\cite{sridhar2024nomad} uses \ac{DDPM}~\cite{chi2023diffusion} for action generation, while FlowNav~\cite{2024flownav} employs \ac{CFM}~\cite{lipman_flow_2022}.  
NaviDiffusor~\cite{zeng2025navidiffusor} is built on top of NoMaD by guiding the diffusion policy to the collision-free space. 
Diffusion policies~\cite{chi2023diffusion} sample from a Gaussian distribution and iteratively denoise the sample using a learned noise-prediction network $\varepsilon_\theta(A_i^k, \mathbf{c}_i, k)$.
The time $k$ indicates the current denoising iteration, while $A_i^k$ denotes the action after $k$ rounds of noise addition. 
Starting from $A_i^1$, drawn from the Gaussian prior, the denoising update is performed as
\begin{equation}
\begin{split}
A_i^{k-1}=\; &\eta(A^k_i-\gamma\varepsilon_\theta\big(A_i^k, \mathbf{c}_i, k)+\mathcal{N}(0,\sigma^2I)\big).
\end{split}
\end{equation}
During training, $k$ is randomly selected to get the noise 
$\varepsilon_k$ 
which is then added to the groundtruth sample $A_i^0$ to obtain $A_i^k$. 
The model is trained by reducing
\begin{equation}
\label{eq:loss-ddpm}
    \mathcal{L} = \text{MSE}\Big(\varepsilon_k, \varepsilon_\theta\big(A_i^k, \mathbf{c}_i, k)\Big).
\end{equation}

\subsection{Action Normalization}

To ensure stable training, the trajectories of length $T_A$ are normalized in two steps before applying the loss in Eq.~\ref{eq:loss-ddpm}. 
The first step compensates for length differences across datasets, since outdoor vehicles typically move faster than indoor robots, as in
\begin{equation}
\label{eq:scale-definition}
\phi_\text{gt} = \frac{\added[id=d]{\sum_{k=0}^{T_A-1}}\lVert\mathbf{a}_{i+(k+1)}-\mathbf{a}_{i+k}\rVert_2}{T_A -1}.
\end{equation}
where $\phi_\text{gt}$ is the average distance between consecutive waypoints, and $\tilde{A}_i = A_i / \phi_\text{gt}$.
We define $\phi_\text{gt}$ as the ground truth waypoint distance.
As in diffusion policy~\cite{chi2023diffusion}, ${\tilde{A}}$ is converted into a sequence of deltas $\Delta \tilde{A}_i$ that represent the displacement between consecutive waypoints,
with ${\Delta  \tilde{\mathbf{a}}_t =  \tilde{\mathbf{a}}_t - \tilde{\mathbf{a}}_{t-1}}$. 
The trajectory is then normalized between $[-1,1]$ using
\begin{equation}
\label{eq:action-stats-norm}
\Delta\hat{\mathbf{a}} =  \frac{ 2\big(\Delta\tilde{\mathbf{a}}-[x_\text{min},y_\text{min}]\big)}{[x_\text{max},y_\text{max}]-[x_\text{min},y_\text{min}]} -1,
\end{equation}
The minima and maxima are calculated across all training datasets.
During inference, $\Delta\tilde{\mathbf{a}}$ is reversed as
\begin{equation}
\Delta\tilde{\mathbf{a}} = \frac{[x_\text{max},y_\text{max}]-[x_\text{min},y_\text{min}]}{ 2\big(\Delta\hat{\mathbf{a}}-[x_\text{min},y_\text{min}]\big)} +1.
\end{equation}
We retrieve the full waypoint trajectory by integrating 
$\tilde{A}_i$ with ${\tilde{\mathbf{a}}_t = \tilde{\mathbf{a}}_{t-1} + \Delta \tilde{\mathbf{a}}_t}$ and ${\tilde{\mathbf{a}}_i=[0,0]}$. 

Note that, during inference, the first normalization step cannot be reversed directly because the \added[id=d]{metric scale to the} real-world \deleted{scale} $\phi_\text{gt}$ is unknown.
To recover the \replaced[id=d]{metric}{real-world} trajectory, previous works apply a handcrafted scaling factor $\phi_\text{tuned}$ as 
\begin{equation}
\label{eq:tuned}
A_i=\phi_\text{tuned}\tilde{A}_i.
\end{equation}

\subsection{Velocity Control}
Earlier approaches deploy the trajectory ${A}_i$ moving directly to the target waypoint $T_W$.
We call this approach velocity control.
The linear velocity $v_i$ and angular velocity $\omega_i$ are defined as 
\begin{equation}
\begin{split}
\label{eq:vel-controller}
    v_i&=\frac{\lVert\mathbf{a}_{i+T_W}\rVert_{2}}{\delta t}\text{ and }\\
    \omega_i&=\frac{\arctan(\tilde{y}_{i+T_W}/\tilde{x}_{i+T_W})}{\delta t},
\end{split}
\end{equation}
where $\tilde{\mathbf{a}}_{i+T_W}=[\tilde{x}_{i+T_W},\tilde{y}_{i+T_W}]$, timestep $\delta t=1/f$, and $f$ is the control frequency. 
This operation assumes that the robot moves to the goal waypoint at $T_W$ in one time unit. 
However, in practice, the robot’s velocity is clipped at $(v_\text{max}, \omega_\text{max})$. 
Since the waypoint $\tilde{\mathbf{a}}_{i+T_W}$ is still normalized by the average waypoint distance, $v_i\gg v_{max}$. 
For instance, in the case of NoMaD~\cite{sridhar2024nomad} where $v_\text{max}=0.4$m/s and $f=15$Hz, if $\tilde{\mathbf{a}}\approx 1$, $v_i=15$m/s for $\phi_\text{tuned}=1$.
Therefore, the state-of-the-art uses $\phi_\text{tuned}=v_\text{max}/f$ in Eq.~\ref{eq:tuned}, which is not necessarily metric scaling.
As a result, $\phi_\text{tuned}$ does not represent the distance between each waypoint in metric space, $\phi_\text{gt}$.
Also, as shown in Fig.~\ref{fig:velocity-collision}, velocity control targets a single waypoint $T_W$ without executing the rest of the planned path, causing the robot to cut corners and collide with obstacles.
The novel addition in our work is learning $\phi_\text{gt}$, allowing trajectories to be projected to the metric space without tuning a scale.

\section{Methodology}

In this section, we first describe our novel network that grounds sampled trajectories from generative policies in \replaced[id=d]{metric}{real-world} coordinates. 
We then define how to act upon a sequence of these grounded waypoints. 
Finally, we show how \replaced[id=d]{metric grounded}{real-world} waypoints can be used to improve the obstacle-avoidance and goal-following capability of diffusion policies.

\subsection{\acs{SCALENET}}

\begin{figure*}[!tb]
\vspace{5.1pt}
    \includegraphics[width=\linewidth]{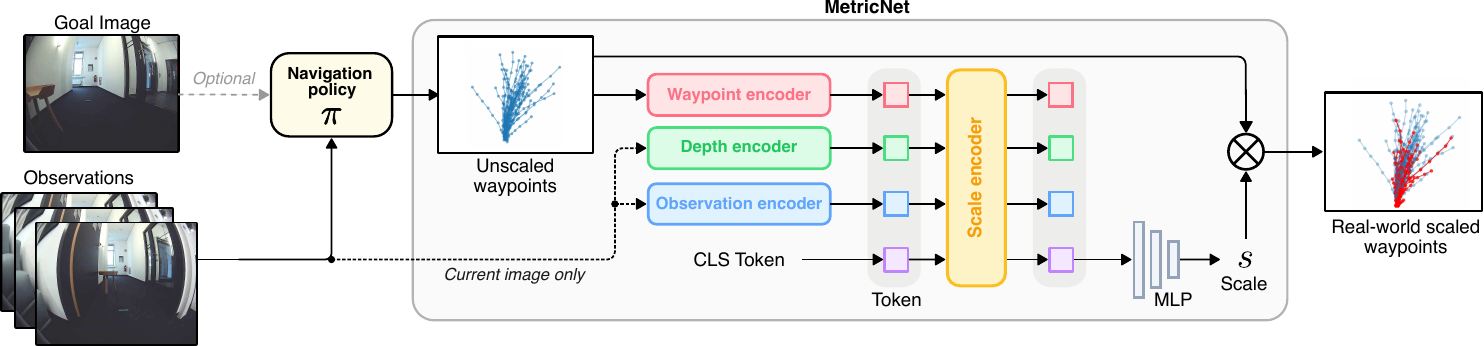}
    \caption{
    \acs{SCALENET} architecture. 
    Our network estimates a factor that converts the unscaled output of generative visual-goal navigation policies into \replaced[id=d]{metric}{real-world} scale. 
    \acs{SCALENET} first tokenizes patches of the current observation using an image encoder. 
    In parallel, the observation is processed with the pre-trained encoder from the Depth-Anything-V2~\cite{yang2025depth} to produce depth patch tokens. 
    A transformer then processes the combined token sequence, prepending a \ac{CLS} token to summarize an aggregated representation. 
    Finally, the output \ac{CLS} token is passed through an MLP to predict the estimated scale.
    This scale is multiplied by the original unscaled waypoints (blue), generating grounded trajectories in \replaced[id=d]{metric}{real-world} coordinates (red).
    }
    \label{fig:scale-scheme}
\end{figure*}

Our architecture is shown in Fig.~\ref{fig:scale-scheme}. 
We encode the unscaled trajectory $\tilde{A}$ using a one-dimensional convolutional network, where $w_\theta(A_i)\in\mathbb{R}^D$ denotes the waypoint encoding with $D=384$.
We split the current observation $\mathbf{o}_i$ into $P_f$ patches of $7\times7$ pixels and encode them using an EfficientNet-B0~\cite{tan2019efficientnet} denoted by $\rho_\theta(\mathbf{o}_i)\in\mathbb{R}^{P_f\times F}$ with $F=1280$. 
These patch embeddings are projected into a common latent dimension $D$ using a residual network.

Like~\cite{2024flownav}, we add depth information to our tokens using the pre-trained DINOv2~\cite{oquab2024dinov2} \added{ViTS} encoder from Depth-Anything-V2~\cite{yang2025depth} to generate patch tokens $d_\theta(\mathbf{o}_i)\in\mathbb{R}^{P_d\times D}$. 
The patch tokens of size $16 \times 16$ pixels are then processed by a single-block residual network, which adds flexibility to the tokens since the DINOv2 encoder remains frozen.

The next block is a transformer encoder $F$ that concatenates the observation, depth, and waypoint tokens using self-attention. 
We prepend a classifier token~$\text{CLS}_\text{in}\in\mathbb{R}^D$ to the beginning of the sequence to capture contextual information~\cite{devlin2019bert}. 
We use a small transformer of six heads with three attention layers each to create the  $\text{CLS}_\text{out}$ token 
\begin{equation}
\text{CLS}_\text{out} = F\big(\text{CLS}_\text{in}, w_\theta(A_i), \rho_\theta(\mathbf{o}_i), d_\theta(\mathbf{o}_i)\big).
\end{equation}
$\text{CLS}_\text{out}$ is then passed through an MLP $g_\theta(\text{CLS}_\text{out})$ to predict a scale $\phi_\text{pred}$ that grounds the waypoints to the metric space.

The objective used to train \acs{SCALENET} is defined as
\begin{equation}
\mathcal{L}(\alpha) = 
        \lambda \cdot \text{MSE}\Big(\phi_\text{pred} , \phi_\text{gt} \Big),
\end{equation} 
where $\lambda$ is a loss scaling factor. 
To avoid gradient collapse, we scale the loss to millimeters ($\lambda=\text{1e-3}$).

\subsection{Position Control}
With \acs{SCALENET}, the trajectory is scaled to metric space, allowing us to move across the entire sequence of waypoints.
We refer to this method as position control.
Since the robot’s view may be occluded (e.g., in curves) or obstacles may be too distant to recognize, we follow only part of the trajectory horizon $T_A$ before recomputing the trajectory~\cite{black2025real}. 
As in velocity control, we set waypoint $T_W$ as the trajectory limit.
However, unlike velocity control, a position controller executes the full segment from $0$ to $T_W$, preventing cutting corners and reducing collisions.

\subsection{\acs{ours}}
\begin{figure*}[!t]
\vspace{5.1pt}
\begin{center}
    \includegraphics[width=.7\linewidth]{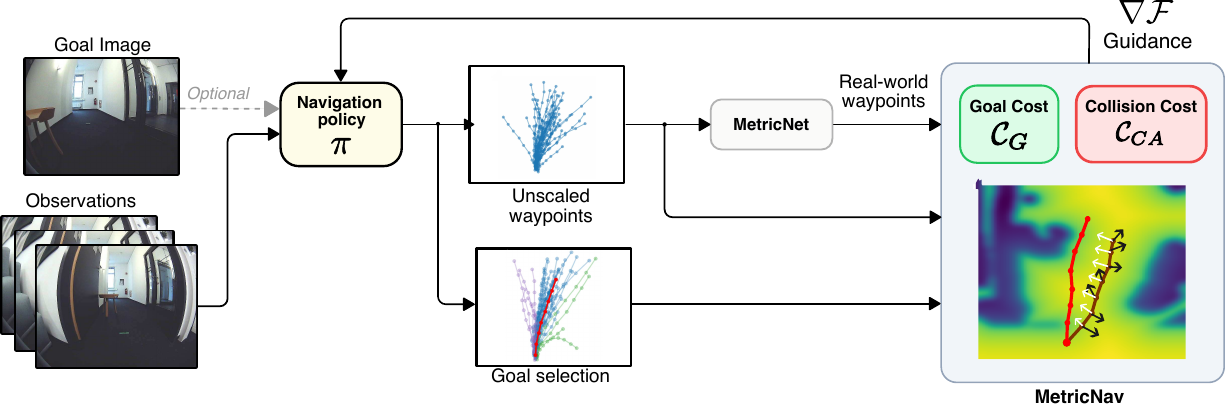}
    \caption{
    \acs{ours} architecture uses a navigation policy to sample $\mathcal{K}$ trajectories (blue).
    Next, spherical k-means clustering is used to estimate the most common goal trajectory (red).
    This trajectory is then used to derive a cost composed of a goal following and a collision avoidance term using an estimated TSDF from monocular depth.
    This cost is then fed into the diffusion policy to guide the sampled trajectory into the obstacle-free space, while moving towards the goal.
    The white arrows show the goal guiding gradients whereas the brown arrows show the collision avoiding gradients.
    }
    \label{fig:METRICNAVREFERENCE}
\end{center}
\end{figure*}

To give the robot additional collision avoidance and goal guidance capabilities, we draw inspiration from previous ideas~\cite{zeng2025navidiffusor,roth2024viplanner} to create \acs{ours}, which uses \acs{SCALENET} to guide any navigation diffusion policy away from obstacles and towards the goal as shown in Fig.~\ref{fig:METRICNAVREFERENCE}.
We define the collision and goal costs which are used for the guiding the diffusion policy and the post diffusion goal selection below.

\subsubsection{Collision Cost}
Given the metric waypoints, we can now project the \replaced[id=d]{rescaled}{real-world} trajectories in a metric pointcloud.
To estimate this pointcloud, we use a metric depth model for monocular images.
Specifically, we use the ViTB variant of Depth-Anything-V2~\cite{yang2025depth}, which is a state-of-the-art model for metric depth estimation.
With approximate intrinsic calibration parameters, the depth image can be converted into an ego-centric pointcloud. 
To make a clear distinction between obstacle and free points, we perform RANSAC-based ground plane segmentation on the pointcloud. 
The ground plane points are marked as free points, while the rest of the points are considered obstacle points.

We use these points to create a local \ac{TSDF}, where the positions of the obstacles correspond to their actual positions in the 3D space. 
In contrast to the \ac{TSDF} representation previously used~\cite{zeng2025navidiffusor,roth2024viplanner}, where non-zero values exist only in the obstacle space, our TSDF representation has non-zero values in both the obstacle and free space. 
This ensures that gradients can still be accumulated to guide trajectories away from the obstacles, even though the sampled paths lie entirely in free space.
The collision cost is defined as $\mathcal{F}_\text{coll}$
\begin{equation}
\begin{split}
    \mathcal{F}_\text{coll}\text{}(A_i^k, \mathbf{o}_i) = \sum_{m=i}^{i+T_A} \big[\mathcal{C}(\mathbf{a}_m) &+\mathcal{C}(\mathbf{a}_m+\sigma_\text{L}) \\
    &+\mathcal{C}(\mathbf{a}_m+\sigma_\text{R})\big],
\end{split}
\label{eq:collision_cost}
\end{equation}
where $\mathcal{C}(\mathbf{a}_m)$ is the cost of each waypoint, and $\mathcal{C}(\mathbf{a}_m+\sigma_\text{L})$ and $\mathcal{C}(\mathbf{a}_m+\sigma_\text{R})$ are the costs at the points that represent the boundaries of the robot.  

\subsubsection{Goal Cost}
Different from NaviDiffusor~\cite{zeng2025navidiffusor}, our cost is explicitly based on the most likely goal direction.
Since diffusion-based navigation policies can generate multiple trajectories that move towards a goal image, we initially sample $\mathcal{K}$ trajectories from the policy.
For an ideal navigation policy, the most likely action would point the robot to the image goal. 
Assuming this to be the case for our base policy, we apply spherical k-means~\cite{hornik2012spherical} to cluster the direction of the $N$-th waypoint of the sampled $\mathcal{K}$ actions. 
The action closest to the cluster center with the largest number of associated actions is chosen as the goal direction.

After selecting a goal, we run a second loop of action generation to define the goal cost function for each scaled action as
\begin{equation}
    \mathcal{F}_{\text{goal}}(A_i) = 1 - \frac{\mathbf{v}_{\text{goal}} \cdot \mathbf{v}_{\mathbf{a}_{i+N}}}{\| \mathbf{v}_{\text{goal}} \|_2 \| \mathbf{v}_{\mathbf{a}_{i+N}} \|_2},
    \label{eq:goal_cost}
\end{equation}
where $\mathbf{v}_{\text{goal}}$ is defined as the direction that points to the goal and ${\mathbf{v}_{\mathbf{a}_i}}$ is defined as the direction of each action based on its $N$-th waypoint.

\subsubsection{Guidance}
Obstacle avoidance is carried out by moving the waypoints towards the free space.
The goal guidance is carried out by maximizing the cosine-similarity between the goal position and the $N$-th waypoints of the generated actions.
We define the total cost as
\begin{equation}
    \mathcal{F}_\text{total}(A_i^k, \mathbf{o}_i) = \alpha \mathcal{F}_\text{goal}(A_i^k) + \beta \mathcal{F}_\text{coll}(A_i^k, \mathbf{o}_i).
\end{equation}
We accumulate gradients for each waypoint in a trajectory by performing a backward pass on this loss function.
The cost $\mathcal{F}(A_i^k, \mathbf{o}_i)$ is then used to guide each denoising step
\begin{equation}
\begin{split}
A_i^{k-1}=\; &\eta(A^k_i-\gamma\varepsilon_\theta\big(A_i^k, \mathbf{c}_i, k)+\mathcal{N}(0,\sigma^2I)\big)
\\&-\mu\nabla_{A_i^k}\mathcal{F}_\text{total}(A_i^k, \mathbf{o}_i).
\end{split}
\end{equation}

By performing a gradient descent on the unnormalized waypoints with step size $\mu$, the sampled trajectories are guided away from collision spaces, while also keeping them pointed towards the general goal direction. We run the guidance steps in diffusion only for $k<2$.
In our experiments, we use $\alpha$ = 0.5, $\beta$ = 0.01 and $\mu$ = 0.1, which were tuned to provide the best goal-guided and collision-free actions. 

\subsubsection{Final Action Selection}
With the use of goal and collision guidance, the actions are safer while also being sufficiently goal-directed.
Given $\mathcal{K}$ sampled actions, we perform an action selection step which includes computing the goal similarity and collision cost for each guided trajectory.
The goal factor $\mathcal{C}_{G}$ is defined as the cosine similarity between an action and the chosen goal.
Collision factor $\mathcal{C}_{CA}$ is defined as the negative of the collision cost in Eq.~\ref{eq:collision_cost}. 
The total cost is defined as
\begin{equation}
    \mathcal{C}_\text{total} = \gamma \mathcal{C}_{G} + (1 - \gamma) \mathcal{C}_{CA}.
\end{equation}
The parameter $\gamma$ pushes trajectories either towards goal guidance or collision avoidance.
The action with the best $\mathcal{C}_\text{total}$ is executed.

\section{Experiments}

\subsection{Training}

We trained three baseline models: 
NoMaD~\cite{sridhar2024nomad}, FlowNav~\cite{2024flownav} and NaviDiffusor~\cite{zeng2025navidiffusor}.
Similar to previous policies, we use a prediction horizon of $T_A=8$ and a goal waypoint $T_W=3$ for both position and velocity control.
We benchmark against the baselines in both real-world and simulation. 
To ensure fairness, all baseline models are trained on the same data split. 
Policies are trained on synthetic data for benchmarking and real-world data for deployment to maximize navigation performance and reduce the sim-to-real gap.
All navigation baselines were trained using the original authors' official implementations and training strategies.
\acs{SCALENET} was trained for 15 epochs on an NVIDIA H200 GPU with a batch size of 1024.
We use AdamW~\cite{loshchilov2019decoupled} optimizer with an initial learning rate of 1e-4 with the cosine scheduler~\cite{loshchilov2022sgdr}.
We train \acs{SCALENET} on both real-world and synthetic data, totaling around 1.5 million data points.

\subsection{Datasets\label{sec:datasets}}

For the real-world data, we combine GoStanford~\cite{hirose2018gonet}, RECON~\cite{shah2021rapid}, Tartan Drive~\cite{triest2022tartandrive}, SACSoN~\cite{hirose2023sacson}, and SCAND~\cite{karnan2022socially}, which contain indoor and outdoor trajectories featuring a diverse range of embodiments and camera parameters.
Similar to NoMaD~\cite{sridhar2024nomad}, we randomly select trajectory segments with a past observation context of length $T_O=4$.
The segments are then normalized following Eq.~\ref{eq:scale-definition} and Eq.~\ref{eq:action-stats-norm} to allow for training on cross-embodied datasets.

We generate synthetic training data by using the Habitat simulator~\cite{savva2019habitat} to collect 16k trajectories across 800 Matterport3D indoor scenes~\cite{chang2017matterport3D}.
We uniformly select camera parameters such as the fisheye angle between 0$^\circ$ and 180$^\circ$ and the camera height between 25cm and 1m.
Trajectories involving stairs are excluded, since our robot is incapable of reproducing this behavior.

For \acs{SCALENET}, the ground truth scale $\phi_\text{gt}$ is collected by averaging the distance between the waypoints in each selected segment.
To increase diversity of $\phi_\text{gt}$, we subsample trajectories by randomly skipping between 1 and 5 waypoints in indoor datasets, and up to 2 waypoints in outdoor datasets.
We subsample synthetic trajectories by spacing them every 1 to 10 waypoints.
While \acs{SCALENET} is trained on synthetic datasets with one observation every 10 cm, the synthetic dataset for policy training uses a coarser resolution, with one observation every 25 cm, as smaller waypoint spacing tended to bias models toward straight-line predictions.

\subsection{Simulation Benchmarking}
For benchmarking in simulation, we compare the baselines over 20 topological maps in 20 different environments. 
\added{A topological map is an ordered collection of images as in ~\cite{sridhar2024nomad,shah2023gnm,shah2022viking}.}
The trajectories are randomly sampled and have a minimum length of 7.5m.
To make a fair comparison, the agents are spawned on the first position of each topological map~\cite{shah2021rapid} and run with the same seed across all baselines.
We test over 5 different seeds, resulting in an evaluation set of 2k repetitions per baseline. 
For collision checking, we use the dimensions of the TurtleBot4.

\subsection{Real-world Deployment}

For real-world experiments, we deploy all models on a TurtleBot4 using ROS2 Humble on Discovery Server mode. 
We use an NVIDIA RTX A500 laptop GPU for model inference.
Following the baselines~\cite{sridhar2024nomad,2024flownav}, we use a fisheye camera capturing images at 15Hz.
To follow the grounded metric waypoints, we use a rotate–translate strategy: the robot first rotates at angular velocity $\omega_{\max}$ for $\arctan(\tilde{y}_{i+T_W})/\omega_{\max}$ seconds, then translates at linear velocity $v_{\max}$ for $\lVert \tilde{\mathbf{a}}_i \rVert/v_{\max}$ seconds.
We avoid using reactive controllers such as the Dynamic Window Approach~\cite{fox2002dynamic} to prevent giving the models unfair advantages.
The test environment is a long corridor \added{of length approximately 20m} with multiple turns and obstacles positioned both at the corners and along the straight segments. 
Each model is evaluated over 5 runs.
We report the average success rate and collision count.
An evaluation is terminated if the robot experiences a frontal collision or a situation where the robot cannot recover within 10 seconds. 

\subsection{Experimental Results}
For navigation, we collect a topological map and measure performance as the maximum percentage of the topomap that the robot can reach before collision.
We follow a similar strategy in the real-world experiments for goal-conditioned navigation.
Exploration experiments are only run in simulation, where we score the distance traveled before collision.

\begin{figure}[!t]

\vspace{5.1pt}
    \includegraphics[width=\linewidth]{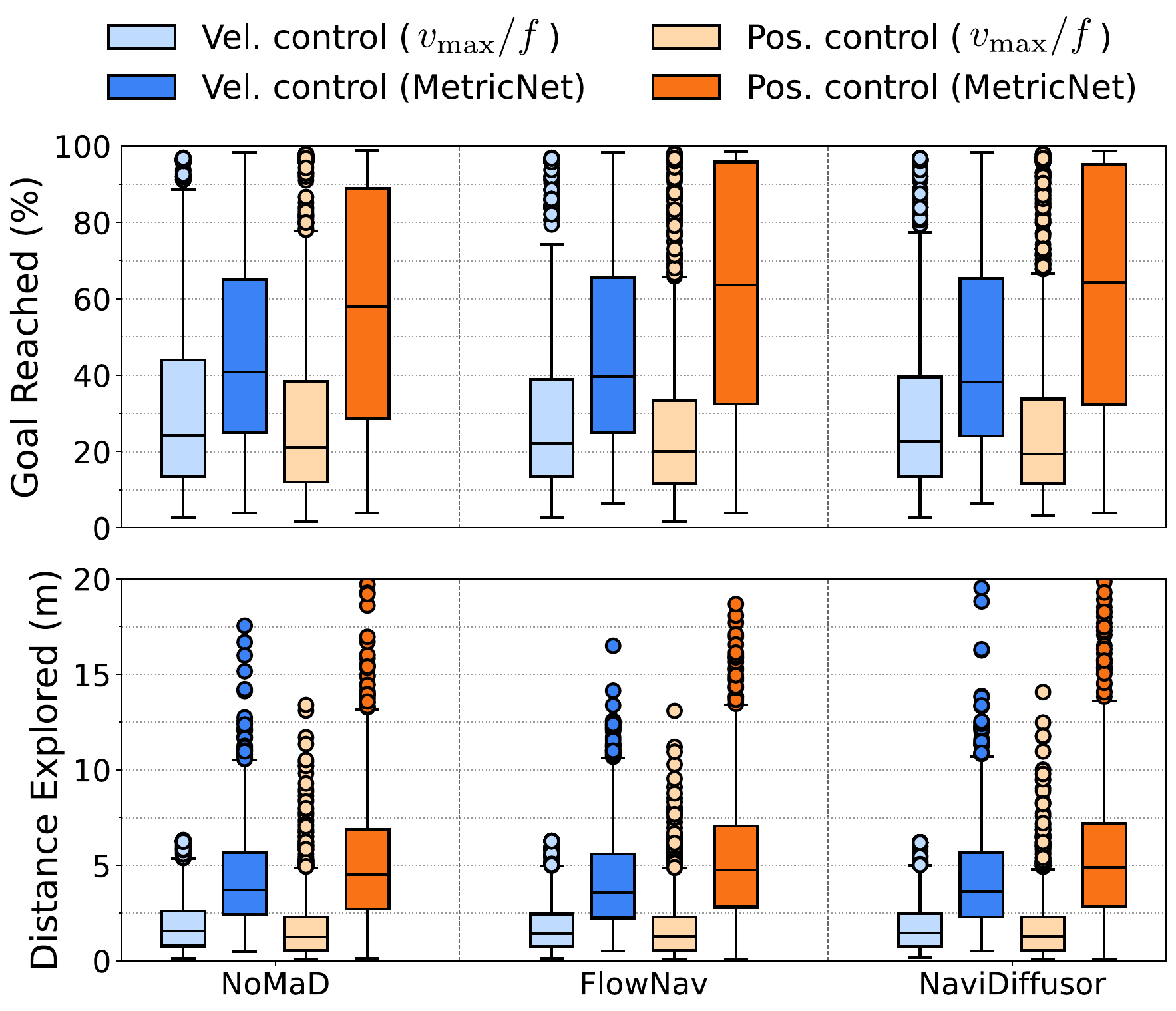}
    \caption{
    Box plot for navigation and exploration in simulation across velocity and position control using a constant scale and prediction by \ac{SCALENET}.
    Using \acs{SCALENET} outperforms using the constant scale proposed by previous work across all base navigation policies. 
    Moreover, using metric waypoints predicted by \acs{SCALENET} with position controller improves the result compared to the velocity control.
    Note that in all navigation experiments, at least one seed fails and one reaches the goal across all baseline policies.
    The key difference lies in the median of the distribution and in how many average seeds succeed.
    }
    \label{fig:pos_vs_vel_ctrl}
\end{figure}

\subsubsection{Comparing different controller paradigms}
Fig.~\ref{fig:pos_vs_vel_ctrl} compares models moving across all waypoints until $T_W$ (position control) or targeting a single waypoint (velocity control). 
In all cases, position control achieves higher success rates and greater explored distance when using \acs{SCALENET} compared to velocity control.
We attribute this improvement to \acs{SCALENET}, which grounds waypoint predictions in metric space, enabling the robot to follow waypoints directly rather than taking incremental steps.
Additionally, we notice that using a fixed scale of $v_{\text{max}}/f$ degrades performance across all models compared to \acs{SCALENET}, highlighting the effectiveness of learning a waypoint scaling factor for better control.
While \acs{SCALENET} improved all baselines, the performance differences between the base policies are smaller than what is reported in other papers. 
This is a direct result of our benchmark, which we designed to be more challenging to test the limits of current models. 
Our environments include tight curves and topological maps created with shortest paths that force the robot to navigate near walls. 
By providing this difficult evaluation, we aim to encourage the development of more robust navigation policies.

\begin{table}[tb]

\vspace{5.2pt}
    \centering
    \caption{Success rate and no. of collisions per run for navigation in real-world deployment}
    \renewcommand{\arraystretch}{1.1}
    \setlength{\tabcolsep}{9pt}
    \begin{tabular}{l|c|c|c|c}
        \toprule
        \multirow{2}{*}{\textbf{Method}} & \multicolumn{2}{c|}{\textbf{Velocity control}} & \multicolumn{2}{c}{\textbf{Position control}} \\
         & SR $\uparrow$ & \#Coll $\downarrow$ & SR $\uparrow$ & \#Coll $\downarrow$ \\ \hline
        NoMaD~\cite{sridhar2024nomad} & 0.76 & 1.0 & 0.87 & 1.8 \\
        FlowNav~\cite{2024flownav}& 0.85 & 1.0 & 0.93 & 1.2\\ 
        NaviDiffusor~\cite{zeng2025navidiffusor} & 0.89 & 0.8 & 0.93 & 0.8 \\ \hdashline
        \rowcolor[HTML]{cffafe} MetricNav (ours) & - & - & \textbf{0.96} & \textbf{0.6}\\
        \bottomrule
    \end{tabular}
    \label{table:real-world}
\end{table}

\begin{table}[tb]
\caption{Average fraction of the topomap completed using position and velocity control in the simulation benchmark}
\label{tab:navigation-simulation}
\centering
\setlength{\tabcolsep}{5pt} 
\renewcommand{\arraystretch}{1.3}
\begin{center}
\begin{tabular}{c|cccc}
\toprule
\multirow{2}{*}{\textbf{Method}} &
  \multicolumn{4}{c}{\textbf{Goal reached $\uparrow$}} \\
 &
  \multicolumn{1}{c|}{NoMaD} &
  \multicolumn{1}{c|}{FlowNav} &
  \multicolumn{1}{c|}{NaviDiffusor} &
  \cellcolor[HTML]{cffafe} MetricNav \\ \hline
\textbf{Vel. control} &
  \multicolumn{1}{c|}{{0.46}} &
  \multicolumn{1}{c|}{{0.45}} &
  \multicolumn{1}{c|}{{0.46}} &
  \cellcolor[HTML]{cffafe} \textbf{0.51} \\ \hline
\textbf{Pos. control} &
  \multicolumn{1}{c|}{0.59} &
  \multicolumn{1}{c|}{\textbf{0.61}} &
  \multicolumn{1}{c|}{\textbf{0.61}} &
  \cellcolor[HTML]{cffafe} \textbf{0.61} \\ \bottomrule
\end{tabular}
\end{center} 
\centering
\end{table}

\subsubsection{Real world deployment}
Real-world navigation results are reported in Table~\ref{table:real-world}, with each model tested using both velocity and position controllers.
Note that position controller is scaled with \acs{SCALENET}, while velocity control uses $v_\text{max}/f$.
Since inference runs on a laptop, \acs{ours} is slower and less suited to velocity control. 
Note that \acs{ours} achieves the highest goal-reaching rate with fewer collisions per run. 
Moreover, all models show improved goal-reaching when paired with \acs{SCALENET} and the position controller compared to velocity control.

\begin{figure*}[t]

\vspace{5.1pt}
    \centering
    \includegraphics[width=\linewidth]{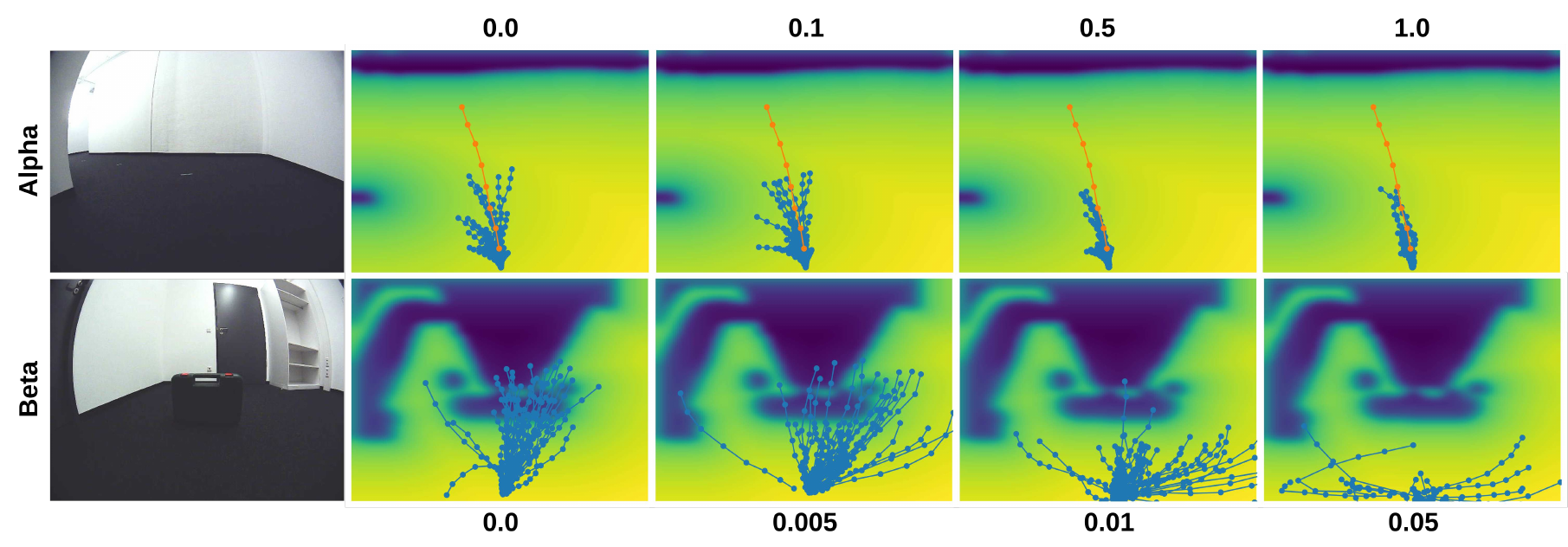} 
    \caption{Example of trajectories generated with different values of $\alpha$ and $\beta$. 
    The larger the value of $\alpha$, the stronger the guidance of the denoised trajectories to the goal.
    The larger the value of $\beta$, the stronger the guidance away from obstacles.
    }
    \label{fig:alpha_beta_variation}
\end{figure*}

\begin{figure}[t]
    \includegraphics[width=\linewidth]{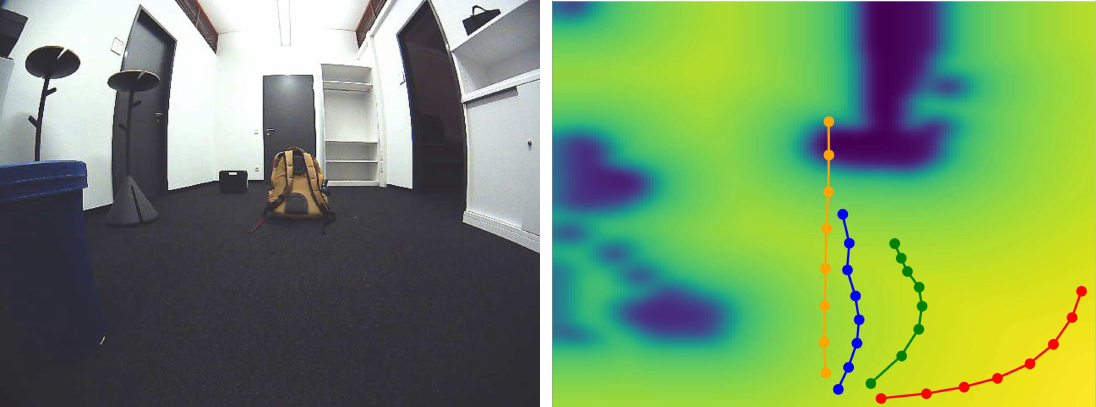} 
    \caption{Example of trajectories produced with different $\gamma$.
    The original selected goal trajectory (orange) collides with the object, whereas all guided and scaled trajectories by \acs{ours} lie in the free space.
    The trajectory with $\gamma=1$ (blue) considers only goal cost and stays close to the goal.
    The trajectory with $\gamma=0$ (red) considers only collision avoidance and prefers driving itself away to a safe state.
    A balance can be found with $\gamma=0.5$ (green).
    }
    \label{fig:gamma_values}
\end{figure}

\subsubsection{Using goal-guidance and metric TSDF}
Table~\ref{tab:navigation-simulation} reports the navigation results for all baselines and \acs{ours} using \replaced[id=d]{metric grounded}{real-world} waypoints produced by \acs{SCALENET}.
Since we do not have the real-world constraints on time synchronization, we benchmark \acs{ours} also in velocity control.
We use $T_W=1$ as a goal waypoint to create a fair comparison to the baselines that do not have active collision avoidance.
Note that \acs{ours} has higher scores in velocity control and comparable results in position control.
\acs{ours} performs better than other baselines in velocity control because it avoids obstacles while still moving towards the goal.

\subsubsection{Qualitative Results}
In Fig.~\ref{fig:alpha_beta_variation} we show how the values chosen for $\alpha$ and $\beta$ for the cost function affect the distribution of guided trajectories that are sampled from the generative policy.
The orange trajectory represents the chosen goal based on the spherical k-means clustering carried out on the $\mathcal{K}$ sampled trajectories.
As $\alpha$ controls the goal guidance, we observe that increasing $\alpha$ values guide the sampled trajectories towards the goal with increasing strength.
As $\beta$ controls the collision avoidance strength, we observe that increasing $\beta$ values moves trajectories away from the obstacles.
In Fig.~\ref{fig:gamma_values} we also show how $\gamma$ affects the chosen action that is executed on the robot.
We fix $\alpha = 0.5$ and $\beta = 0.01$ and plot actions for $\gamma \in \{0, 0.5, 1\}$.
The orange trajectory represents the chosen goal trajectory, the blue trajectory ($\gamma=1.0$) closely follows it, the red trajectory ($\gamma=0.0$) prioritizes collision avoidance, and the green trajectory ($\gamma=0.5$) balances both.
This shows that unguided trajectories often cause collisions, whereas guided ones help the robot progress towards the goal while avoiding obstacles.

\section{Conclusions}
In this paper, we present \acs{SCALENET}, a network that predicts the scale factor to project unnormalized trajectories from navigation diffusion policies into metric space.
Grounding trajectories in \replaced[id=d]{metric}{real-world} space avoids hand-tuning calibration and allows multiple waypoints to be executed, significantly improving performance in both navigation and exploration. 
We demonstrate that using metric waypoints from \acs{SCALENET} in our guidance framework \acs{ours} helps diffusion navigation policies avoid obstacles and move towards predefined image goals.
We validate our method in simulation and real-world experiments.
As future work, we propose using a controller that allows for reactiveness to dynamic obstacles while executing the metric waypoints, such as the Dynamic Window Approach.

\section*{Acknowledgments}
 The authors gratefully acknowledge the scientific support and HPC resources provided by the Erlangen National High Performance Computing Center (NHR@FAU) of the Friedrich-Alexander-Universität Erlangen-Nürnberg (FAU) under the BayernKI project v106be. BayernKI funding is provided by Bavarian state authorities. We also acknowledge the support by the Körber European Science Prize. This work has been partially supported by the German Federal Ministry of Research, Technology and Space (BMFTR) under the Robotics Institute Germany (RIG).

\bibliographystyle{IEEEtran}
\bibliography{root}

\end{document}